\documentclass[10pt,twocolumn,letterpaper]{article}

\usepackage{cvpr}
\usepackage{times}
\usepackage{epsfig}
\usepackage{graphicx}
\usepackage{amsmath}
\usepackage{amssymb}
\usepackage{textcomp}
\usepackage{subcaption}
\usepackage{cite}

\usepackage{float}
\usepackage{longtable,booktabs}
\usepackage{algorithm}
\usepackage[noend]{algpseudocode}
\usepackage{multirow}

\usepackage[pagebackref=true,breaklinks=true,letterpaper=true,colorlinks,bookmarks=false]{hyperref}

\renewcommand{\vec}[1]{\boldsymbol{#1}}

\DeclareMathOperator*{\argmin}{arg\,min}

\cvprfinalcopy 
\def\cvprPaperID{474} 

\usepackage{fancyhdr}
\fancypagestyle{plain}{%
 \fancyhf{}
\chead{\textit{To appear in the 2019 IEEE Conference on Computer Vision and Pattern Recognition (CVPR)}}
\cfoot{1}
}

\begin{document}
    
    \title{End-to-end Projector Photometric Compensation}
	\author{Bingyao Huang$^{1,2,}$\thanks{Work partly done during internship with HiScene.} \;\;\;\;\; Haibin Ling$^{1,}$\thanks{Corresponding author.}\\
		{\normalsize $^1$Department of Computer and Information Sciences, Temple University, Philadelphia, PA USA}\\
		{\normalsize $^2$Meitu HiScene Lab, HiScene Information Technologies, Shanghai, China}\\
		{\tt\small \{bingyao.huang, hbling\}temple.edu}
	}
	
	\maketitle

	\begin{abstract}
		Projector photometric compensation aims to modify a projector input image such that it can compensate for disturbance from the appearance of projection surface. 
		%
		In this paper, for the first time, we formulate the compensation problem as an end-to-end learning problem and propose a convolutional neural network, named \emph{CompenNet}, to implicitly learn the complex compensation function.
		CompenNet consists of a UNet-like backbone network and an autoencoder subnet. Such architecture  encourages rich multi-level interactions between the camera-captured projection surface image and the input image, and thus captures both photometric and environment information of the projection surface. In addition, the visual details and interaction information are carried to deeper layers along the multi-level skip convolution layers. The architecture is of particular importance for the projector compensation task, for which only a small training dataset is allowed in practice.
		
		Another contribution we make is a novel evaluation benchmark, which is independent of system setup and thus quantitatively verifiable. Such benchmark is not previously available, to our best knowledge, due to the fact that conventional evaluation requests the hardware system to actually project the final results. Our key idea, motivated from our end-to-end problem formulation, is to use a reasonable surrogate to avoid such projection process so as to be setup-independent.
		Our method is evaluated carefully on the benchmark, and the results show that our end-to-end learning solution outperforms state-of-the-arts both qualitatively and quantitatively by a significant margin.
	\end{abstract}
	
	\section{Introduction}
	
	\begin{figure}[!t]
		\begin{center}
			\includegraphics[width=1.0\linewidth]{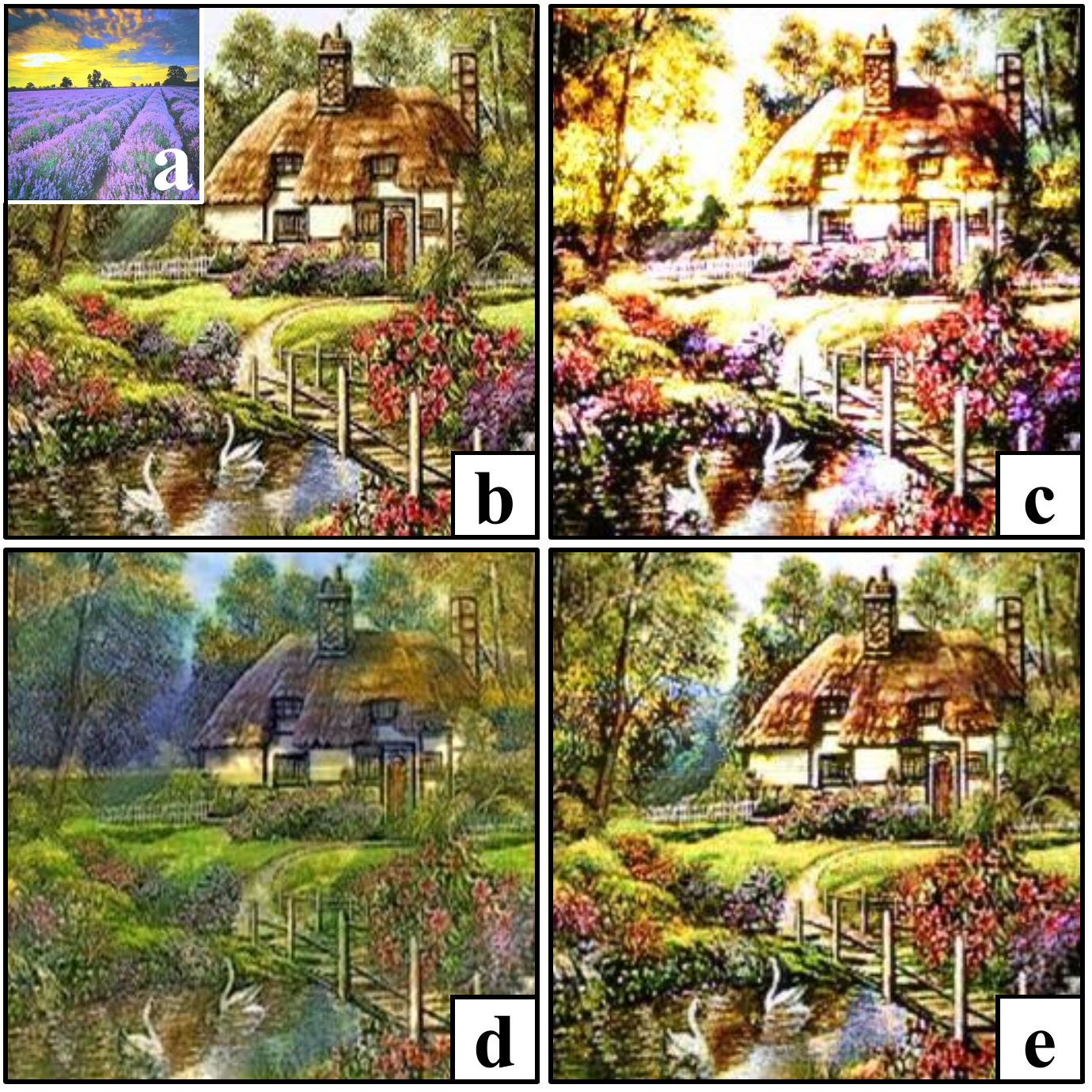}
			\vspace{-5mm}\caption{Projector photometric compensation. \textbf{(a)} Textured projection surface under normal illumination. \textbf{(b)} Input image (desired visual effect). \textbf{(c)} Camera-captured uncompensated projection result, \ie, (b) projected onto (a). \textbf{(d)} Compensated image by the proposed CompenNet. \textbf{(e)} Camera-captured compensated projection result, \ie, (d) projected onto (a). Comparing (c) and (e) we can see clearly improved color and details. }\label{fig:tisser}
		\end{center}
		\vspace{-6mm}
	\end{figure}
	
	Projectors are widely used in applications such as presentation, cinema, structured light and projection mapping \cite{Geng2011,raskar2003ilamps, yoshida2003virtual,grossberg2004making, bimber2005embedded, aliaga2012fast,siegl2015real, siegl2017adaptive, narita2017dynamic}. To ensure high perception quality, existing systems typically request the projection surface (screen) to be white and textureless, under reasonable environment illumination. Such request, however, largely limits applicability of these systems.
	Projector photometric compensation~\cite{bimber2005embedded, yoshida2003virtual, aliaga2012fast,raskar2003ilamps, siegl2017adaptive, siegl2015real, narita2017dynamic}, or simply \emph{compensation} for short, aims to address this issue by modifying a projector input image to compensate for the projection surface as well as associated photometric environment. An example from our solution is illustrated in Fig.~\ref{fig:tisser}, where the compensated projection result (e) is clearly more visually pleasant than the uncompensated one (c).
	
	A typical projector compensation system consists of a camera-projector pair and a projection surface placed at a fixed distance and orientation. Firstly, the projector projects a sequence of sampling input images to the projection surface, then the sampling images are absorbed, reflected or refracted according to the projection surface material. Once the camera captures all the projected sampling images, a composite radiometric transfer function is fitted that maps the input images to the captured images. This function (or its inverse) is then used to infer the compensated image for a new input image.
	Existing solutions (\eg, \cite{nayar2003projection,grossberg2004making, sajadi2010adict,grundhofer2015robust}) usually model the compensation function explicitly, with various simplification assumptions that allow the parameters to be estimated from samples collected. These assumptions, such as context independence (\S\ref{sec:related_works}), however, are often violated in practice. Moreover, due to the tremendous complexity of the photometric process during projection, reflection and capturing, it is extremely hard, if not impossible, to faithfully model the compensation explicitly.
	
	In this paper, for the first time, an end-to-end projector compensation solution is presented to address the above issues. We start by reformulating the compensation problem to a novel form that can be learned online, as required by the compensation task in practice. This formation allows us to develop a convolutional neural network (CNN), named \emph{CompenNet}, to implicitly learn the complex compensation function.
	In particular, CompenNet consists of two subnets, a UNet-like \cite{ronneberger2015u} backbone network and an autoencoder subnet. Firstly, the autoencoder subnet encourages rich multi-level interactions between the camera-captured projection surface image and the input image, and thus captures both photometric and environment information of the projection surface. Secondly, the UNet-like backbone network allows the visual details and interaction information to be carried to deeper layers and the output using the multi-level skip convolution layers. The two subnets together make CompenNet efficient in practice and allow CompenNet to learn the complex backward mapping from camera captured image to projector input image. 	In addition, a pre-trained solution is designed that can further improve the training efficiency with a small tradeoff in precision.
	
	Another issue addressed in this paper is the absence of evaluation benchmarks for projector compensation, due mainly to the fact that traditional evaluation is highly setup dependent. More specifically, to evaluate a compensation algorithm, theoretically, its experimental results need to be actually projected and captured and then quantitatively compared with ground truth. This process makes it impractical to provide a shared benchmark among different research groups. In this work, we tackle this issue by deriving a surrogate evaluation protocol that requests no actual projection of the algorithm output. As a result, this surrogate allows us to construct, for the first time, a sharable setup-independent compensation benchmark.

	
	
	The proposed compensation network, \ie, CompenNet, is evaluated on the proposed benchmark that is carefully designed to cover various challenging factors. In the experiments, CompenNet demonstrates clear advantages compared with state-of-the-art solutions.
	In summary, in this paper we bring the following contributions:
	\vspace{-.5em}
	\begin{enumerate}
		\setlength{\itemsep}{0pt}
		\setlength{\parsep}{1pt}
		\setlength{\parskip}{1pt}
		\item For the first time, an end-to-end solution is proposed for projector compensation. Such solution allows our system to effectively and implicitly capture the complex photometric process involved in the projector compensation process.
		\item The proposed CompenNet is designed to have two important subnets that enable rich multi-level interactions between projection surface and input image, and to carry interaction information and structural details through the network.
		\item A pre-train method is proposed to further improve the practical efficiency of our system.
		\item For the first time, a setup-independent projector compensation benchmark is constructed, which is expected to facilitate future works in this direction.
	\end{enumerate}
	\vspace{-2mm}The source code, benchmark and experimental results are available at {\small\url{https://github.com/BingyaoHuang/CompenNet}}.

	\section{Related Works}\label{sec:related_works}
	In theory, the projector compensation process is a very complicated nonlinear function involving the camera and the projector sensor radiometric responses \cite{li2018practical}, lens distortion/vignetting \cite{juang2007photometric}, defocus \cite{zhang2006projection, yang2016practical}, surface material reflectance and inter-reflection \cite{takeda2016inter}. A great amount of effort has been dedicated to designing practical and accurate compensation models, which can be roughly categorized into context-independent \cite{nayar2003projection, grossberg2004making, sajadi2010adict,  grundhofer2015robust} and context-aware ones \cite{ashdown2006robust, aliaga2012fast, takeda2016inter, li2018practical}. Detailed reviews can be found in \cite{bimber2008visual, grundhofer2018recent}.

	\vspace{.5mm}\noindent\textbf{Context-independent methods} typically assume that there is an approximate one-to-one mapping between the projector and camera image pixels, \ie, a camera pixel is only dependent on its corresponding projector pixel and the surface patch illuminated by that projector pixel. Namely, each pixel is roughly independent of its neighborhood context.
	The pioneer work by Nayar \etal \cite{nayar2003projection} proposes a linear model that maps a projector ray brightness to camera detected irradiance with a 3$\times$3 color mixing matrix. Grossberg \etal \cite{grossberg2004making} improve Nayar's work and model the environment lighting by adding a 3$\times$1 vector to the camera-captured irradiance. However, a spectroradiometer is required to calibrate the uniform camera radiometric response function. Moreover, as pointed out in \cite{juang2007photometric}, even with a spectroradiometer the assumption of uniform radiometric response is usually violated, let alone the linearity.
	Considering the nonlinearity of the transfer function, Sajadi \etal \cite{sajadi2010adict} fit a smooth higher-dimensional B{\'e}zier patches-based model with 9\textsuperscript{3}=729 sampling images. Grundh{\"o}fer and Iwai \cite{grundhofer2015robust} propose a thin plate spline (TPS)-based method and reduce the number of sampling images to 5\textsuperscript{3}=125 and further deal with clipping errors and image smoothness with a global optimization step.
	Other than optimizing the image colors numerically, some methods specifically focus on human perceptual properties, \eg Huang \etal \cite{huang2017radiometric} generate visually pleasing projections by exploring human visual system's chromatic adaptation and perceptual anchoring property. Also, clipping artifacts due to camera/projector sensor limitation are minimized using gamut scaling.
	
	Despite largely simplifying the compensation problem, the context-independent assumption is usually violated in practice, due to many factors such as projector distance-to-surface, lens distortion, defocus and surface inter-reflection \cite{zhang2006projection, yang2016practical,takeda2016inter}. Moreover, it is clear that a projector ray can illuminate multiple surface patches, a patch can be illuminated by the inter-reflection of its surrounding patches, and a camera pixel is also determined by rays reflected by multiple patches. 
	
	\vspace{.5mm}\noindent\textbf{Context-aware methods} compensate a pixel by considering information from neighborhood context. Grundh{\"o}fer \etal \cite{grundhofer2008real} tackle visual artifacts and enhance brightness and contrast by analyzing the projection surface and input image prior. Li \etal \cite{li2018practical} reduce the number of sampling images to at least two by sparse sampling and linear interpolation. Multidimensional reflectance vectors are extracted as color transfer function control points.
	Due to the small size of sampling dots, this method may be sensitive to projector defocus and lens vignetting. A simple linear interpolation using those unreliable samples may add to the compensation errors. Besides computing an offline compensation model, Aliaga \etal \cite{aliaga2012fast} introduce a run time linear scaling operation to optimize multiple projector compensation.  Takeda \etal \cite{takeda2016inter} propose an inter-reflection compensation method using an ultraviolet LED array.
	
	Context-aware methods generally improve over previous methods by integrating more information. However, it is extremely hard to model or approximate the ideal compensation process due to complex interactions between the global lighting, the projection surface and the input image. Moreover, most existing works focus on reducing pixel-wise color errors rather than jointly improve the color and structural similarity to the target image.
	
	Our method belongs to the Context-aware one, and in fact captures much richer context information by using the CNN architecture. Being the first end-to-end learning-based solution, our method implicitly and effectively models the complex compensation process. Moreover, the proposed benchmark is the first one that can be easily shared for verifiable quantitative evaluation.
	
	\vspace{1mm}\noindent\textbf{Our method} is inspired by the successes of recently proposed \textbf{deep learning-based image-to-image translation}, such as pix2pix \cite{isola2017image}, CycleGAN \cite{zhu2017unpaired}, style transfer \cite{johnson2016perceptual, gatys2016image, huang2017arbitrary}, image super-resolution \cite{dong2014learning, kim2016accurate, ledig2017photo, wang2018sftgan} and image colorization \cite{zhang2016colorful, iizuka2016let, deshpande2017learning}. That said, as the first deep learning-based projector compensation algorithm, our method is very different from these studies and has its own special constraints. For example, unlike above CNN models that can be trained once and for all, the projector compensation model needs to be quickly retrained if the system setup changes. However, in practice, both capturing training images and training the model are time consuming. In addition, data augmentations such as cropping and affine translations are not available for our task, because each camera pixel is strongly coupled with a neighborhood of its corresponding projector pixel and the projection surface patch illuminated by those pixels. Furthermore, general image-to-image translation models cannot formulate the complex spectral interactions between the global lighting, the projector backlight and the projection surface. In fact, in our evaluation, the advantage of the proposed method over the classical pix2pix \cite{isola2017image} algorithm is clearly demonstrated quantitatively and qualitatively.

	\begin{figure*}[!t]
		\begin{center}
			\includegraphics[width=1\linewidth]{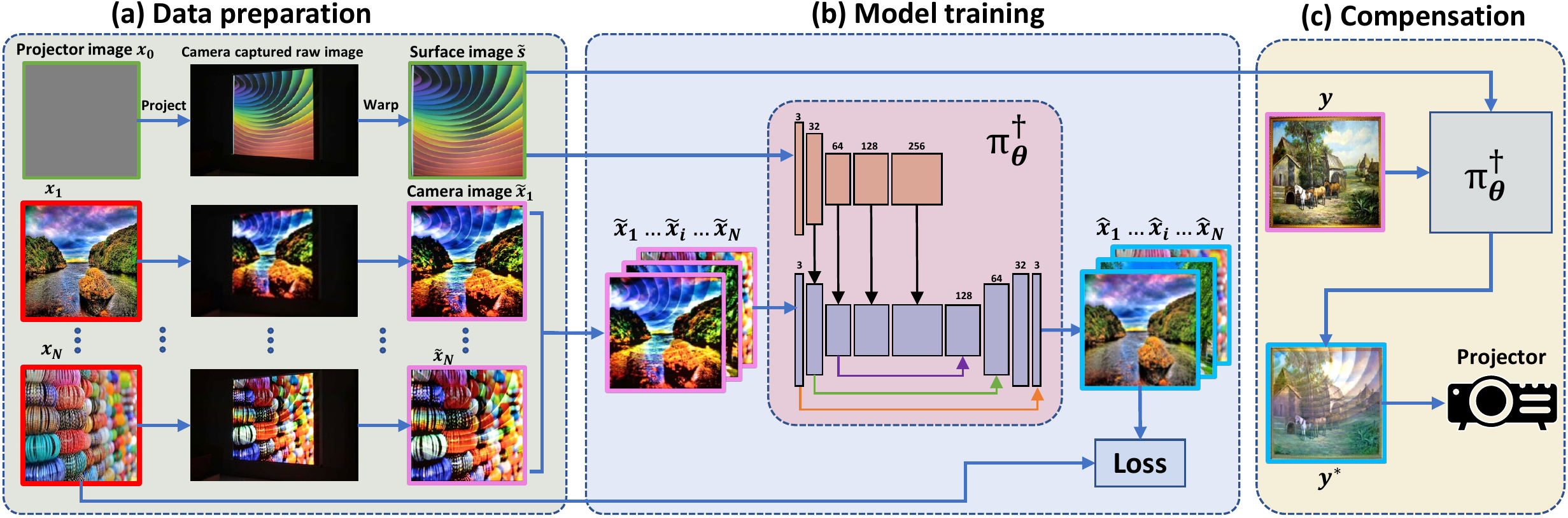}
			\vspace{-2mm}\caption{Flowchart of the proposed projector compensation pipeline consisting of three major steps. \textbf{(a)} Project and capture a surface image and a set of sampling images. \textbf{(b)} The proposed CompenNet, \ie, $\pi^{\dagger}_{\vec{\theta}}$, is trained using the projected and captured image pairs. \textbf{(c)} With the trained model, an input image $\vec{y}$ can be compensated and projected.}   \label{fig:flowchart}
			\vspace{-6mm}		\end{center}
	\end{figure*}
	\section{Deep Projector Compensation}\label{sec:problem}
	\subsection{Problem formulation}\label{subsec:problem_formulation}
	Our projector compensation system consists of a camera-projector pair and a planar projection surface placed at a fixed distance and orientation. Let a projector input image be $\vec{x}$; and let the projector's and the camera's composite geometric projection and radiometric transfer functions be $ \pi_p $ and $ \pi_c $, respectively. Let the surface spectral reflectance property and spectral reflectance functions be $\vec{s}$ and $\pi_s$,  respectively. Let the global lighting irradiance distribution be $\vec{g}$, then the camera captured image $\tilde{\vec{x}}$,\footnote{We use `tilde' ($\tilde{\vec{x}}$) to indicate a camera-captured image.} is given by:
	\begin{equation}\label{eq:cam}
	\tilde{\vec{x}} = \pi_c\Big(\pi_s\big(\pi_p( \vec{x} ), \vec{g}, \vec{s}\big)\Big)
	\end{equation}
	
	
	The problem of projector compensation is to find a projector input image $ \vec{x}^{*}$, named \emph{compensation image} of $ \vec{x} $ such that the camera captured image is the same as the ideal desired viewer perceived image, \ie,
	\begin{equation}\label{eq:cmp_rad}
	\pi_c\Big(\pi_s\big(\pi_p( \vec{x}^{*} ), \vec{g}, \vec{s}\big)\Big) = \vec{x}
	\end{equation}
	
	However, the spectral interactions and responses formulated in the above equation are very complex and can hardly be solved by traditional methods. Moreover, in practice it is also hard to measure $ \vec{g} $ and $ \vec{s} $ directly. For this reason, we capture their spectral interactions using a camera-captured surface image $\tilde{\vec{s}} $ under the global lighting and the projector backlight:
	\begin{equation}\label{eq:g_s}
	\tilde{\vec{s}} = \pi_c\Big(\pi_s\big(\pi_p(\vec{x}_0), \vec{g}, \vec{s}\big)\Big),
	\end{equation}
	where $\vec{x}_0$ is theoretically a black image. In practice, the projector outputs some backlight $\pi_p(\vec{x}_0)$ even when the input image is black, thus we encapsulate this factor in $\tilde{\vec{s}}$. When under low global illumination, $\tilde{\vec{s}} $ suffers from camera gamut clipping due to the limitation of camera dynamic range,  thus we set $\vec{x}_0$ to a plain gray image to provide some illumination. 	Denoting the composite projector to camera radiometric transfer function in Eq.~\ref{eq:cmp_rad} as $\pi$ and substituting $\vec{g}$ and $\vec{s}$ with $\tilde{\vec{s}}$, we have the compensation problem as
	\begin{equation}\label{eq:cmp_composite2}
	\pi( \vec{x}^{*}; \tilde{\vec{s}}) = \vec{x} \ \  \Rightarrow \ \ \vec{x}^{*} = \pi^{\dagger}(\vec{x}; \tilde{\vec{s}}),
	\end{equation}
	where $ \pi^{\dagger} $ is the pseudo-inverse function of $ \pi $ and, obviously, has no closed form solution.

	\subsection{Learning-based formulation}
	A key requirement for learning-based solution is the availability of training data. In the following we derive a method for collecting such data.
	Investigating the formulation in \S\ref{subsec:problem_formulation} we find that:
	\begin{equation}\label{eq:final}
	\tilde{\vec{x}} = \pi(\vec{x}; \tilde{\vec{s}}) \ \ \Rightarrow \ \ \vec{x} = \pi^{\dagger}(\tilde{\vec{x}}; \tilde{\vec{s}})
	\end{equation}
	This suggests that we can learn $\pi^{\dagger}$ over sampled image pairs like $(\tilde{\vec{x}}, \vec{x})$ and a surface image $\tilde{\vec{s}}$ as shown in Fig.~\ref{fig:net}.
	In fact, some previous solutions (\eg \cite{sajadi2010adict, grundhofer2015robust}) use similar ideas to fit models for $\pi^{\dagger}$, but typically under simplified assumptions and without modeling $\tilde{\vec{s}}$.
	
	Instead, we reformulate the compensation problem with a deep neural network solution, which is capable of preserving the projector compensation complexity. In particular, we model the compensation process with an end-to-end learnable convolutional neural network, named \emph{CompenNet} and denoted as $\pi^{\dagger}_{\vec{\theta}}$ (see (Fig.~\ref{fig:flowchart}(b)), such that
	\begin{equation}\label{eq:cnn_pred}
	\vec{\hat{x}} = \pi^{\dagger}_{\vec{\theta}}(\tilde{\vec{x}}; \tilde{\vec{s}}),
	\end{equation}
	where $ \vec{\hat{x}} $ is the compensation of $\tilde{\vec{x}}$ (not $\vec{x}$) and $ \vec{\theta} $ contains the learnable network parameters. It is worth noting that $ \tilde{\vec{s}} $ is fixed as long as the the setup is unchanged, thus only one  $ \tilde{\vec{s}} $ is needed in training and prediction.
	
	By using Eq.~\ref{eq:final}, we can generate a set of $N$ training pairs, denoted as $\mathcal{X}=\{(\tilde{\vec{x}}_i, \vec{x}_i)\}_{i=1}^N $. Then, with a loss function $ \mathcal{L}$, CompenNet can be learned by
	\begin{equation}\label{eq:objective}
	\vec{\theta} = \argmin_{\vec{\theta}'}\sum_i\mathcal{L}\big(\vec{\hat{x}}_i=\pi^{\dagger}_{\vec{\theta}'}(\tilde{\vec{x}_i}; \tilde{\vec{s}}), \ \vec{x}_i\big)
	\end{equation}
	Our loss function is designed to jointly optimize the compensated image's color and structural similarity to the target image by combining the pixel-wise $ \ell_1 $ and the SSIM loss:
	\begin{equation}\label{eq:loss}
	\mathcal{L}  = \mathcal{L}_{\ell_1} + \mathcal{L}_{\text{SSIM}}
	\end{equation}
	The advantages of this loss function over the other loss functions are shown in \cite{zhao2017loss} and in our comprehensive experimental comparisons in Table~\ref{tab:compare_loss} and Fig.~\ref{fig:compare_loss}.
	
	%
	
	\subsection{Network design}\label{subsec:cnn_architecture}
	\begin{figure*}[!t]
		\begin{center}
			\includegraphics[width=1.0\linewidth]{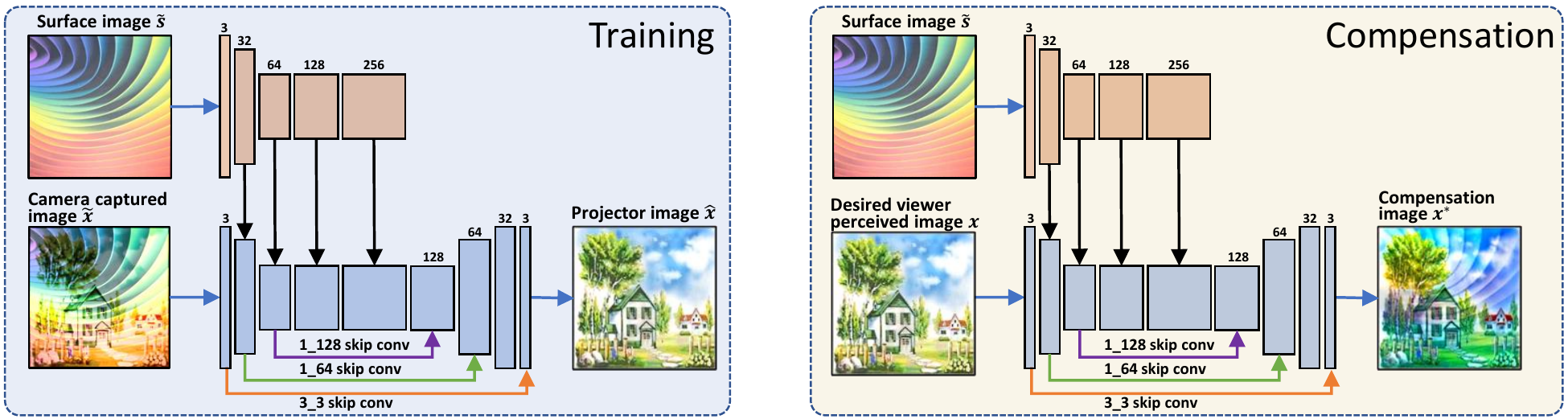}
			\caption{The architecture of CompenNet (ReLU layers omitted). All convolution layers are composed of 3$\times$3 filters and all transposed convolution layers consist of 2$\times$2 filters. Both upsample and downsample layers use a stride of two. The number of filters for each layer is labeled on its top. The skip convolution layers are shown in colored arrows, and the number of layers and the number of filters are labeled as \textbf{\#layers\_\#filters} for conciseness. Learning the backward mapping from camera captured uncompensated image to the projector input image (left: $ \tilde{\vec{x}} \mapsto  \vec{x} $) is the same as learning the mapping from desired viewer perceived image to the compensation image (right: $ \vec{x} \mapsto  \vec{x}^*$).}\label{fig:net}
		\end{center}
		\vspace{-6mm}
	\end{figure*}
	
	Based on the above formulation, our CompenNet is designed with two input images, $\tilde{\vec{x}}$ and $\tilde{\vec{s}}$, corresponding to the camera-captured uncompensated image of $\vec{x}$ and the camera-captured surface image, respectively. The network architecture is shown in Fig.~\ref{fig:net}. Both two inputs and the output are 256$\times$256$\times$3 RGB images. Both input images are fed to a sequence of convolution layers to downsample and to extract multi-level feature maps. Note that in Fig.~\ref{fig:net} we give the two paths different colors to indicate that the two branches do NOT share weights. The multi-level feature maps are then combined by element-wise addition, allowing the model to learn the complex spectral interactions between the global lighting, the projector backlight, the surface and the projected image. 
	
	We also pass low-level interaction information to high-level feature maps through skip convolution layers \cite{he2016deep}. In the middle blocks, we extract rich features by increasing the feature channels while keeping the feature maps' width and height unchanged. Then, we use two transposed convolution layers to gradually upsample the feature maps to 256$\times$256$\times$32. Finally, the output image is an element-wise summation of the last layer's output and the three skip convolution layers at the bottom of Fig.~\ref{fig:net}. Note that we clamp the output image pixel values to [0, 1] before output. We find that deeper CNNs with more layers and filters, \eg 512 filters can produce better compensation results, but suffer from overfitting on fewer sampling images and longer training and prediction time. However, if an application prefers accuracy to speed, it can add more convolution layers, increase the number of iterations and capture more training data accordingly. In this paper, we choose the architecture in Fig.~\ref{fig:net} to balance training/prediction time and sampling data size.
	
	To make the method more practical, we also provide a pre-trained model by projecting and capturing $ N (N = 500) $ sampling images using a white projection surface. Once the setup, \eg, the projection surface or the global lighting changes, rather than recapturing 500 training images, we use much fewer (\eg 32) images to fine-tune the pre-trained model. This technique saves data preparation and training time and adds to the advantages over the existing solutions. We demonstrate the effectiveness of the pre-trained model in \S\ref{subsec:pre-train}.

	\subsection{Training details}\label{subsec:cnn_training}
	We implement CompenNet using PyTorch \cite{paszke2017automatic} and train it using Adam optimizer \cite{kinga2015method} with the following specifications: we set $ \beta_{1} = 0.9$ and fix $\ell_2 $ penalty factor to $ 10^{-4} $. The initial learning rate is set to $ 10^{-3} $, we also decay it by a factor of 5 every 800 iterations. The model weights are initialized using He's method \cite{he2015delving}. We train the model for 1,000 iterations on two Nvidia GeForce 1080 GPUs with a batch size of 64, and it takes about 10min to finish training (for 500 samples). We report a comprehensive evaluation of different hyperparameters in supplementary material.

	\subsection{Compensation pipeline}
	To summarize, the proposed projector compensation pipeline consists of three major steps shown in Fig.~\ref{fig:flowchart}.  \textbf{(a)} We start by projecting a plain gray image $ \vec{x}_0 $ and $ N $ sampling images $ \vec{x}_1, \dots, \vec{x}_N $ to the planar projection surface and capture them using the camera. Then each captured image is warped to the canonical view using a homography, and we denote warped camera images as $ \tilde{\vec{x}}_i $. \textbf{(b)} Afterwards, we gather the $ N $ image pairs $ (\tilde{\vec{x}}_i, \vec{x}_i) $ and train the compensation model $ \pi^{\dagger}_{\vec{\theta}} $. \textbf{(c)} Finally, with the trained model, we generate the compensation image $ \vec{y}^{*} $ for an input image $ \vec{y} $ and project $ \vec{y}^{*} $ to the surface.
	
	\section{Benchmark}\label{sec:implementation}
	
	An issue left unaddressed in previous studies is the lack of public benchmarks for quantitative evaluation, due mainly to the fact that traditional evaluation is highly setup-dependent. In theory, to evaluate a compensation algorithm, its output compensation image $\vec{x}^*$ for input $\vec{x}$ should be actually projected to the projection surface, and then captured by the camera and quantitatively compared with the ground truth. This process is obvious impractical since it requests the same projector-camera-environment setup for fair comparison of different algorithms.
	
	In this work, motivated by our problem formulation, we derive an effective surrogate evaluation protocol that requests no actual projection of the algorithm output. The basic idea is, according to Eq.~\ref{eq:final}, we can collect testing samples in the same way as the training samples. We can also evaluate an algorithm in the similar way.
	Specifically, we collect the test set of $M$ samples as $\mathcal{Y}=\{(\tilde{\vec{y}}_i, \vec{y}_i)\}_{i=1}^M$, under the same system setup as the training set $\mathcal{X}$. Then the algorithm performance can be measured by averaging over similarities between each test input image $\vec{y}_i$ and its algorithm output $\hat{\vec{y}}_i = \pi^{\dagger}_{\vec{\theta}}(\tilde{\vec{y}}_i; \tilde{\vec{s}})$.
	
	The above protocol allows us to construct a projector compensation evaluation benchmark, consisting of $K$ system setups, each with a training set $\mathcal{X}_k$, a test set $\mathcal{Y}_k$ and a surface image $\tilde{\vec{s}}_k$, $k=1,\dots,K$.
	
	\vspace{.5mm}\noindent\textbf{System configuration.}
	Our projector compensation system consists of a Canon 6D camera with image resolution of 960$\times$640, and a ViewSonic PJD7828HDL DLP projector set to the resolution of 800$\times$600. The distance between the camera and the projector is 500mm and the projection surface is around 1,000mm in front of the camera-projector pair. The camera exposure mode, focus mode and white balance mode are set to manual, the global lighting is fixed during the data capturing and system validation.
	
	\vspace{.5mm}\noindent\textbf{Dataset.}
	To obtain the sampling colors and textures as diverse as possible, we download 700 colorful textured images from the Internet and use $ N=500 $ for each training set $\mathcal{X}_k$ and $ M = 200$ for each testing set $\mathcal{Y}_k$. In total $K = 24$ different setups are prepared for training and evaluation. Future works can replicate our results and compare with CompenNet on the benchmark without replicating our setups.  For more camera perceived compensation results and the detailed configurations of the benchmark please refer to supplementary material. 
	


	\section{Experimental Evaluations}\label{sec:experiments}
	
	\begin{figure*}[!t]
		\begin{center}
			\includegraphics[width=1\linewidth]{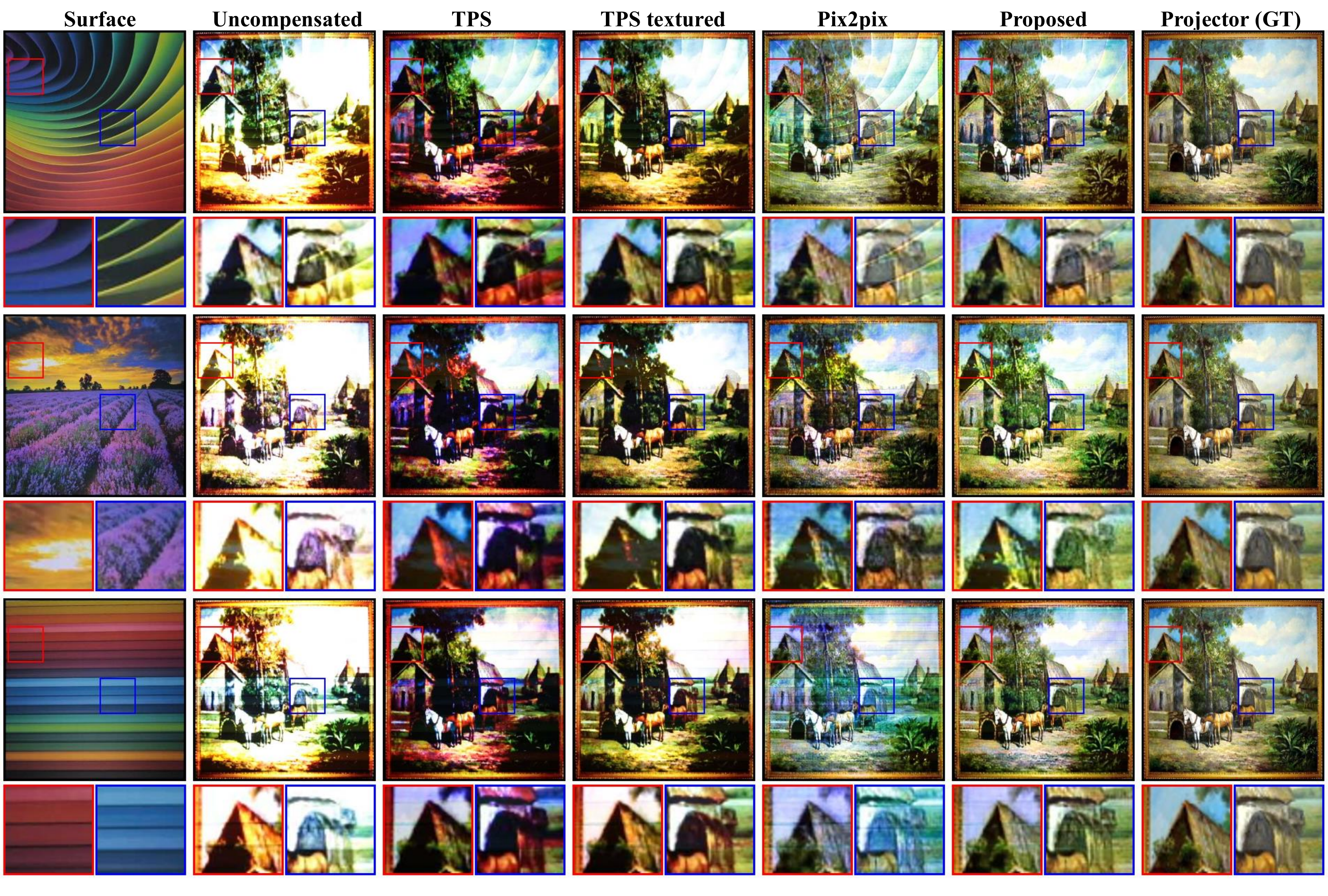}
			\vspace{-6mm}
			\caption{Comparison of TPS \cite{ grundhofer2015robust}, TPS textured, pix2pix \cite{isola2017image} and CompenNet on different surfaces. The 1\textsuperscript{st} column is the camera-captured projection surface. The 2\textsuperscript{nd} column is the camera-captured uncompensated projected image. The 3\textsuperscript{rd} to 6\textsuperscript{th} columns are the camera-captured compensation results of different methods. The last column is the ground truth input image. Each image is provided with two zoomed-in patches for detailed comparison.  When trained with diverse textured images, TPS produces better results than its original version \cite{grundhofer2015robust} that uses plain color images, though still suffers from hard edges, blocky effect and color errors. Compared with CompenNet, pix2pix generates unsmooth pixelated details and color errors. 
			}
			\label{fig:compare_existing}
		\end{center}
		\vspace{-5mm}
	\end{figure*}
	
	\begin{figure*}[!t]
		\begin{center}
			\includegraphics[width=1\linewidth]{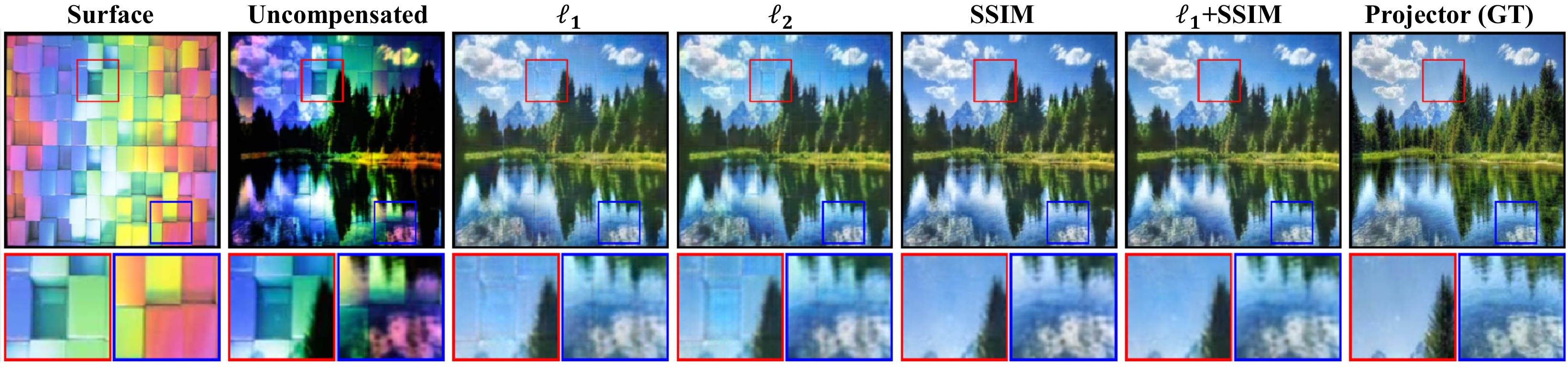}
			\vspace{-5mm}
			\caption{Qualitative comparison of CompenNet trained with $ \ell_1$ loss, $\ell_2$ loss, SSIM loss and $ \ell_1 +$SSIM loss. It shows that the $ \ell_1 $ and $ \ell_2 $ losses are unable to successfully compensate the surface patterns. The $ \ell_1 +$SSIM and the SSIM losses produce similar results, but the water in the zoomed-in patch of SSIM is bluer than the $ \ell_1 +$ SSIM and the ground truth.}\label{fig:compare_loss}
		\end{center}
		\vspace{-6mm}
	\end{figure*}
	
	\subsection{Comparison with state-of-the-arts}\label{subsec:comparison_existing}
	\vspace{-2mm}We compare the proposed projector compensation method with a context-independent TPS model \cite{grundhofer2015robust}, an improved TPS model (explained below) and a general image-to-image translation model pix2pix \cite{isola2017image} on our benchmark.
	
	We first capture 125 pairs of plain color sampling image as used in the original TPS method~\cite{grundhofer2015robust}. We also fit the TPS method using our diverse textured training set $\mathcal{X}_k$, and name this method \textbf{TPS textured}. The experiment results in Table~\ref{tab:compare} and Fig.~\ref{fig:compare_existing} show clear improvement of TPS textured over the original TPS method.
	
	We then compare our method with pix2pix \cite{isola2017image} to demonstrate the challenge of the  projector compensation problem and to show the advantages of our formulation and architecture. We use the default implementation\footnote{{\scriptsize \url{https://github.com/junyanz/pytorch-CycleGAN-and-pix2pix}}} of pix2pix with some adaptations for the compensation problem:
	(1) as mentioned in \S\ref{sec:related_works}, data augmentation can break the strong coupling between the camera, the surface and the projector image, thus, we disable cropping, resizing and flipping. (2) We train the pix2pix model for 10,000 iterations and it takes about 10min with a batch size of one using the same hardware. 	The comparison results show that our method outperforms pix2pix by a significant margin on this task.
	
	We find that TPS textured obtains slightly increased SSIM and slightly decreased PSNR when the data size increases. Pix2pix shows the lowest PSNR and SSIM when training data size is 250, and the highest PSNR and SSIM at 500. Only the proposed CompenNet achieves higher PSNR and SSIM when training data size increases from 125 to 500 (Table~\ref{tab:compare}). Despite improving the performance of the CompenNet, a downside of large data size is increased data capturing time. In practice, taking hundreds of sampling images is time consuming, therefore, we proposed a pre-trained model that has improved performance than the default model when we only have limited training pairs and training time (\S\ref{subsec:pre-train}).
	
	Besides the state-of-the-arts above, we also tested the model-free ``refinement by continuous feedback" method in \cite{nayar2003projection} and find it work well. However, it has the disadvantage of needing several real projections, captures and iterations to converge for \textit{each single} frame. Thus, it is impractical to evaluate it on the proposed setup-independent surrogate evaluation benchmark.
	
	\subsection{Effectiveness of the surface image}\label{subsec:surf}
	To show the effectiveness of our learning-based formulation and that the surface image $ \tilde{\vec{s}} $ is a necessary model input, we compare with the proposed CompenNet that is without the input surface image and the corresponding autoencoder subnet, we name it \textbf{CompenNet w/o surf}. The results are shown in Table~\ref{tab:compare}. Firstly,  we can see a clear increase in PSNR and SSIM and a drop in RMSE when the $ \tilde{\vec{s}} $ is included in the model input (\textbf{CompenNet}). This shows that our learning-based formulation has a clear advantage over the models that ignore the important information encoded in the surface image. Secondly, \textbf{CompenNet w/o surf} outperforms TPS, TPS textured and pix2pix on  PSNR, RMSE and SSIM even $ \tilde{\vec{s}} $ is not included. It is worth noting that for a new projection setting, simply replacing the surface image does not work well and it is necessary to train a new CompenNet from scratch. Fortunately, with the pre-trained model we can fine-tune from a reasonable initialization to reduce the number of training images and training time.
	
	
	\begin{table}
		\begin{center}
			\caption{Quantitative comparison of compensation algorithms. Results are averaged over $K=24 $ different setups.\vspace{-2mm}  
			}\label{tab:compare}
			{\small \begin{tabular}[]{c@{}l@{} lll}
					\toprule
					\textbf{\#Train}~ & \textbf{Model} & \textbf{PSNR}$\uparrow$ & \textbf{RMSE}$\downarrow$ & \textbf{SSIM}$\uparrow$\tabularnewline
					\midrule
					125 & TPS \cite{grundhofer2015robust} & 17.6399 & 0.2299 & 0.6042\tabularnewline & TPS \cite{grundhofer2015robust} textured & 19.3156 & 0.1898 & 0.6518\tabularnewline & Pix2pix \cite{isola2017image} & 17.9358 & 0.2347 & 0.6439\tabularnewline & CompenNet w/o Surf. & 19.8227 & 0.1786 & 0.7003\tabularnewline & CompenNet & \textbf{21.0542} & \textbf{0.1574} & \textbf{0.7314}\tabularnewline
					\midrule
					250 & TPS \cite{grundhofer2015robust} textured & 19.2764 & 0.1907 & 0.6590\tabularnewline & Pix2pix \cite{isola2017image} & 16.2939 & 0.2842 & 0.6393\tabularnewline & CompenNet w/o Surf. & 20.0857 & 0.1733 & 0.7146\tabularnewline & CompenNet & \textbf{21.2991} & \textbf{0.1536} & \textbf{0.7420}\tabularnewline
					\midrule
					500 & TPS \cite{grundhofer2015robust} textured & 19.2264 & 0.1917 & 0.6615\tabularnewline & Pix2pix \cite{isola2017image} & 18.0923 & 0.2350 & 0.6523\tabularnewline & CompenNet w/o Surf.~ & 20.2618 & 0.1698 & 0.7209\tabularnewline & CompenNet & \textbf{21.7998} & \textbf{0.1425} & \textbf{0.7523}\tabularnewline
					\midrule
					- & Uncompensated & 12.1673 & 0.4342 & 0.4875\tabularnewline
					\bottomrule
			\end{tabular}}
		\end{center}
		\vspace{-5mm}
	\end{table}
	
	\subsection{Effectiveness of the pre-trained model}\label{subsec:pre-train}
	\vspace{-2mm}We compare the default CompenNet model (using He's \cite{he2015delving} initialization) with a model that is pre-trained with 500 training pairs projected to a \textbf{white} surface. Then we train and evaluate both models on each training set $\mathcal{X}_k$ and evaluation set $\mathcal{Y}_k$ of the 24 setups that the models have never been trained on. To demonstrate that pre-trained model obtains improved performance with limited training pairs and training time, we train the models for 500 iterations using only 32 training pairs. The results are reported in Table~\ref{tab:compare_pretrain}.
	
	Clearly, we see that the pre-trained model outperforms the default counterpart even the 24 training and evaluation setups have different lightings and surface textures as the pre-trained setup. Our explanation is that despite the surfaces have different appearances, the pre-trained model has already learned partial radiometric transfer functions of the camera and the projector.  This pre-trained model makes our method more practical, \ie, as long as the projector and camera are not changed, the pre-trained model can be quickly tuned with much fewer training images and thus shortens the image capturing and training time. Another finding is even with 32 training pairs and 500 iterations, the proposed CompenNet, with or without pre-train, performs better than TPS \cite{grundhofer2015robust}, TPS textured and pix2pix \cite{isola2017image} in Table~\ref{tab:compare}. Furthermore, CompenNet has much fewer parameters (1M) than pix2pix's default generator (54M parameters). This further confirms that the projector compensation is a complex problem and is different from general image-to-image translation tasks, and carefully designed models are necessary to solve this problem.

	\subsection{Comparison of different loss functions}\label{subsec:loss}
	Existing works fit the composite radiometric transfer function by linear/nonlinear regression subject to a pixel-wise $ \ell_2 $ loss and this loss function is known to penalize large pixel errors while oversmoothes the structural details. We investigate four different loss functions, i.e., pixel-wise $ \ell_1 $ loss, pixel-wise $ \ell_2 $ loss, SSIM loss, and  $ \ell_1 +$SSIM loss. The qualitative and quantitative comparisons are shown in Fig.~\ref{fig:compare_loss} and Table~\ref{tab:compare_loss}, respectively. Compared with SSIM loss, pixel-wise $ \ell_1 $ and $ \ell_2 $ losses cannot well compensate surface patterns, notice the hard edges in the red zoomed-in patches in Fig.~\ref{fig:compare_loss}. Consistent to the qualitative results, the SSIM column in Table~\ref{tab:compare_loss} also shows a clear disadvantage of both pixel-wise $ \ell_1 $ and $ \ell_2 $ losses. Although SSIM loss alone obtains the best SSIM value, its PSNR and RMSE are the 2\textsuperscript{nd} worst. After comprehensive experiments on our benchmark, we find that $ \ell_1 +$SSIM loss obtains the best PSNR/RMSE and the 2\textsuperscript{nd} best SSIM, thus, we choose it as our CompenNet loss function. Moreover, even when trained with pixel-wise $ \ell_1 $ loss, CompenNet outperforms TPS, TPS textured and pix2pix on  PSNR, RMSE and SSIM, this again shows a clear advantage of our task-targeting formulation and architecture.

	\begin{table}
		\begin{center}
			\caption{Quantitative comparison between a pre-trained CompenNet and a CompenNet randomly initialized using He's method \cite{he2015delving}, both trained using \textbf{only 32 samples and 500 iterations} with a batch size of 32, and take about 170s.\vspace{-3mm}
			}\label{tab:compare_pretrain}		
			{\small \begin{tabular}[]{lccc}
					\toprule
					\textbf{Model} & \textbf{PSNR}$\uparrow$ & \textbf{RMSE}$\downarrow$ & \textbf{SSIM}$\uparrow$\tabularnewline
					\midrule
					CompenNet & 19.6767 & 0.1863 & 0.6788\tabularnewline
					CompenNet pre-train & \textbf{20.3423} & \textbf{0.1688} & \textbf{0.7165}\tabularnewline\midrule		
					Uncompensated & 12.1673 & 0.4342 & 0.4875\tabularnewline
					\bottomrule
			\end{tabular}}
		\end{center}
		\vspace{-5mm}
	\end{table}
	
	\begin{table}
		\begin{center}
			\caption{Quantitative comparison of different loss functions for the proposed CompenNet.\vspace{-3mm}
			}\label{tab:compare_loss}		
			{\small \begin{tabular}[]{lccc}
					\toprule
					\textbf{Loss} & \textbf{PSNR}$\uparrow$ & \textbf{RMSE}$\downarrow$ & \textbf{SSIM}$\uparrow$\tabularnewline
					\midrule					
					$ \ell_1 $ & 21.1782 & 0.1527 & 0.6727\tabularnewline
					$ \ell_2 $ & 20.7927 & 0.1594 & 0.6453\tabularnewline
					SSIM & 21.0134 & 0.1566 & \textbf{0.7591}\tabularnewline
					$ \ell_1 +$SSIM & \textbf{21.7998} & \textbf{0.1425} & 0.7523\tabularnewline\midrule
					Uncompensated & 12.1673 & 0.4342 & 0.4875\tabularnewline
					\bottomrule
			\end{tabular}}
		\end{center}
		\vspace{-5mm}
	\end{table}
	
	\subsection{Limitations}\label{subsec:limitation}
	\vspace{-1mm}We focus on introducing the first end-to-end solution to projector compensation, for planar surfaces with decent, not necessarily ideal, reflectance/geometric qualities. In addition, we have not experimented on surfaces with special reflectance transport properties, such as water, strong specular reflection, geometry inter-reflection and semi-gloss, thus it may not work well in these cases.
	
	\section{Conclusions}\label{sec:conclusions}
	\vspace{-2mm}In this paper, we reformulate the projector compensation problem as a learning problem and propose an accurate and practical end-to-end solution named CompenNet. In particular,  CompenNet explicitly captures the complex spectral interactions between the environment, the projection surface and the projector image. The effectiveness of our formulation and architecture is verified by comprehensive evaluations. Moreover, for the first time, we provide the community with a novel setup-independent evaluation benchmark dataset. Our method is evaluated carefully on the benchmark, and the results show that our end-to-end learning solution outperforms state-of-the-arts both qualitatively and quantitatively by a significant margin. To make our model more practical, we propose a pre-train method, which adds to the advantages over the prior works.
	
	\vspace{.5mm}\noindent{\textbf{Acknowledgement.} We thank the anonymous reviewers for valuable and inspiring comments and suggestions. 
	}

	{\small
		\bibliographystyle{ieee_fullname}
		\bibliography{ref}
	}

\end{document}